\documentclass[11pt]{article}

\usepackage[final]{acl}
\usepackage{times}
\usepackage{latexsym}

\usepackage[T1]{fontenc}

\usepackage[utf8]{inputenc}

\usepackage{microtype}

\usepackage{inconsolata}

\usepackage{graphicx}

\usepackage{natbib}
\usepackage{hyperref}
\usepackage{amssymb}
\usepackage{amsmath}
\usepackage{enumitem}
\usepackage{wrapfig}
\usepackage{makecell}
\usepackage{multirow}
\usepackage{subfigure}
\usepackage{subcaption}
\usepackage{algorithm}
\usepackage{algpseudocode}
\usepackage{pifont}
\usepackage{booktabs} 
\usepackage{colortbl}
\usepackage{stfloats}
\setlist[itemize]{leftmargin=10pt}

\title{Context-Agent: Dynamic Discourse Trees for Non-Linear Dialogue}

\author{Junan Hu, Shudan Guo, Wenqi Liu, Jianhua Yin, Yinwei Wei\thanks{Corresponding author.} \\
  Shandong University, China \\
  \texttt{junanhu@mail.sdu.edu.cn, weiyinwei@hotmail.com}}

\begin{document}
\maketitle
\begin{abstract}
  Large Language Models demonstrate outstanding performance in many language tasks but still face fundamental challenges in managing the non-linear flow of human conversation. The prevalent approach of treating dialogue history as a flat, linear sequence is misaligned with the intrinsically hierarchical and branching structure of natural discourse, leading to inefficient context utilization and a loss of coherence during extended interactions involving topic shifts or instruction refinements. To address this limitation, we introduce Context-Agent, a novel framework that models multi-turn dialogue history as a dynamic tree structure. This approach mirrors the inherent non-linearity of conversation, enabling the model to maintain and navigate multiple dialogue branches corresponding to different topics. Furthermore, to facilitate robust evaluation, we introduce the Non-linear Task Multi-turn Dialogue (NTM) benchmark, specifically designed to assess model performance in long-horizon, non-linear scenarios. Our experiments demonstrate that Context-Agent enhances task completion rates and improves token efficiency across various LLMs, underscoring the value of structured context management for complex, dynamic dialogues. The dataset and code is available at \href{https://github.com/Steve2457/Context-Agent}{GitHub}.
\end{abstract}

\section{Introduction}

The advancement of dialogue systems based on LLMs is pivotal for the efficacy of next-generation applications, including AI Agents and collaborative robotics, where the ability to maintain context-aware communication is fundamental to task completion and user engagement \citep{durante2024agent, yao2024tau, sun2026topodim}. Following the advent of LLMs' context window expansion techniques, the capabilities for multi-turn dialogue have been significantly enhanced \citep{li2025beyond}.

\begin{figure}[h]
  \centering
  \includegraphics[width=0.36\textwidth]{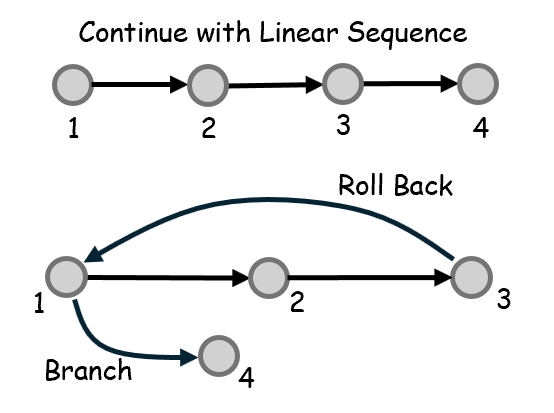}
  \caption{A schematic diagram of linear (upper) vs. non-linear (lower) dialogue flow.}
  \label{fig:simple}
\end{figure}

However, LLMs still grapple with a fundamental challenge inherent to natural human conversation: the management of non-linear dialogue flow. This phenomenon occurs when conversational topics do not advance in a sequential order but instead feature shifts, topical jumps, or interwoven threads of discussion \citep{laban2025llms}. Such non-linear dynamics are commonplace in real-world interactions, where users may revisit previous topics, introduce new subjects, or refine earlier statements based on evolving understanding or context \citep{mann1988rhetorical}. The prevalent approach of treating dialogue history as a flat, linear sequence is fundamentally misaligned with the intrinsic structure of human conversation \citep{wang2024survey, li2025beyond}. This linear paradigm fails to capture the hierarchical and branching nature of dialogues, leading to inefficiencies in context utilization and challenges in maintaining coherence over extended interactions \citep{lian2026sweagilesoftwareagentframework,DBLP:conf/icml/DingZZXSX0Y24}.

Effectively resolving the non-linear flow problem requires overcoming several challenges. The first is the accurate identification and management of topic shifts and instruction refinements within a conversation. The second is the efficient selection of context from a potentially vast and complex dialogue history. As conversations extend over multiple turns, the accumulation of information can lead to increased computational costs and the risk of overwhelming the model with irrelevant details \citep{DBLP:conf/iclr/JorenZFJTR25, jiang2026rlpo}, leading to the ``needle in a haystack'' problem \citep{DBLP:journals/tacl/LiuLHPBPL24, vaswani2017attention}. The third challenge lies in the development of robust evaluation metrics and benchmarks that can accurately assess a model's performance in handling non-linear dialogues, as existing datasets often lack the complexity and variability found in real-world interactions.

To address these challenges, inspired by the hierarchical organization inherent in human cognitive processes for managing complex dialogues \citep{grosz1986attention}, we propose Context-Agent, a novel framework that models multi-turn dialogue history as a dynamic tree. This approach allows for the representation of conversations in a way that reflects their inherent non-linear nature, enabling the model to maintain multiple branches of dialogue corresponding to different topics. Furthermore, recognizing the inadequacy of existing datasets for this problem, we introduce the Non-linear Task Multi-turn Dialogue (NTM) benchmark, specifically designed to evaluate the performance of models in long-horizon, non-linear dialogue scenarios. This benchmark features dialogues with multiple topic shifts and instruction refinements, providing a more realistic and challenging testbed for assessing context management strategies.

In summary, the main contributions of this paper are as follows:
\begin{itemize}
\item We propose \textbf{Context-Agent}, a novel framework that models dialogue history as a dynamic tree. This approach captures non-linear discourse structure, enabling precise context navigation via tree structure.
\item We introduce the \textbf{Non-linear Task Multi-turn Dialogue (NTM)} benchmark. It features long-horizon dialogues with complex topic shifts and instruction refinements, offering a rigorous testbed for non-linear context management.
\item Experiments across various LLMs demonstrate that Context-Agent significantly outperforms linear baselines, improving task completion rates while reducing token usage.
\end{itemize}

\section{Related Works}

\label{gen_inst}

\textbf{Linear Context Extension and Compression.} While recent works have explored structured and task-aware parameter-efficient fine-tuning \citep{xiao2026not}, architectures for context extension like YaRN \citep{DBLP:conf/iclr/PengQFS24} and LongLoRA \citep{DBLP:conf/iclr/ChenQTLL0J24} extend context windows but face high computational costs and the ``lost-in-the-middle'' problem \citep{DBLP:journals/tacl/LiuLHPBPL24}. Conversely, compression methods \citep{su2022speaker, DBLP:conf/emnlp/ParkKY21} reduce token usage but degrade performance by flattening dialogue structure, sacrificing details essential for complex reasoning.

\textbf{Structured Memory and Retrieval.} Retrieval-Augmented Generation (RAG) adapts external retrieval to internal dialogue history, with various methods addressing data quality and mitigating retrieval-induced hallucinations \citep{zhang2026stable, ma2024context}. While flat retrieval methods like DH-RAG \citep{zhang2025dh} filter irrelevant turns, they often retrieve fragmented segments that lack local coherence. Recent advances have moved towards structured memory. Notably, MemTree \citep{rezazadeh2024isolated} and RAPTOR \citep{sarthi2024raptor} organize information into hierarchical tree structures.

\begin{table*}[t]
\centering
\small
\resizebox{\textwidth}{!}{
\begin{tabular}{lccccc}
\toprule[1.5pt]
\textbf{Method} & \textbf{Structure} & \textbf{Construction Basis} & \textbf{Retrieval Unit} & \textbf{Local Coherence} & \textbf{Update Efficiency} \\
\midrule[1pt]
\multicolumn{6}{l}{\textit{Linear \& Compression Methods}} \\
\cmidrule(lr){1-6}
Full Context & Linear Sequence & Token Concatenation & Entire History & High & Very Low ($O(N^2)$) \\
MemGPT & OS-like Hierarchy & Event-Triggered/Function & Paginated Memory & High (Self-Edit) & Medium \\
\midrule[0.5pt]
\multicolumn{6}{l}{\textit{Retrieval-Augmented Generation (RAG)}} \\
\cmidrule(lr){1-6}
Standard RAG & Flat Index & Semantic Similarity & Indep. Chunks & Low (Disjointed) & High \\
DH-RAG & Chain & Semantic Clustering & Query Chains & High (Dynamic) & Medium (Incremental) \\
\midrule[0.5pt]
\multicolumn{6}{l}{\textit{Tree-Structured Memory}} \\
\cmidrule(lr){1-6}
RAPTOR & Static Tree & Bottom-up Clustering & Abstractive Summaries & High & Low (Offline Rebuild) \\
MemTree & Dynamic Tree & Online Clustering & Collapsed Nodes & Medium (Disjointed) & High ($O(\log N)$) \\
\midrule[1pt]
\textbf{Context-Agent (Ours)} & \textbf{Dynamic Tree} & \textbf{Discourse Intent} & \textbf{Coherent Path} & \textbf{Very High (Path-Aware)} & \textbf{High (Event-Triggered)} \\
\bottomrule[1.5pt]
\end{tabular}
}
\caption{Comparison of context management paradigms. We compare our method with linear methods, standard RAG, advanced RAG, and tree-based memory.}
\label{tab:comparison}
\end{table*}

Table \ref{tab:comparison} delineates the distinctions between our framework and existing paradigms. A fundamental limitation of current structured approaches, such as MemTree, lies in their reliance on \textbf{semantic similarity} for aggregation, grouping content based on textual overlap rather than \textbf{discourse flow}. This often conflates distinct conversational threads that share lexical features but diverge in intent. Conversely, \textbf{Context-Agent} explicitly models \textbf{discourse structure} \citep{grosz1986attention}. By constructing trees based on \textbf{navigational intent} (e.g., instruction refinement, topic switching) and retrieving coherent \textbf{paths} instead of isolated nodes, our approach preserves the logical continuity requisite for complex, long-horizon tasks.

\section{Method}

\label{method}

Our framework models a multi-turn dialogue as a forest of topic trees. Each tree represents a distinct topic and is composed of nodes (dialogue units) and branches. The dialogue's evolution is managed through state transitions.

\subsection{Formal Problem Definition}
Conventional dialogue systems model history as a linear sequence $H_t = \{(q_1, r_1), \ldots, (q_t, r_t)\}$, generating a response $r_{t+1}$ from a query $q_{t+1}$ via a function $g(H_t, q_{t+1})$. This flat representation leads to contextual redundancy and loss of structural information.

To address this limitation, we introduce and formalize the problem of Non-linear Contextual Dialogue Management. The central premise of this problem is to shift from treating the entire history $H_t$ as an undifferentiated input to representing it as a dynamically evolving, hierarchically structured dialogue forest, denoted as $F_t$.

We model the interaction flow as a dynamic tree to align with the Attentional State theory \citep{grosz1986attention}. This theory posits that human cognitive focus operates hierarchically, managing a focus stack rather than a connected graph. Explicit graph structures risk violating local coherence by merging distinct branches, thereby introducing noise from competing contexts. In contrast, our tree framework enforces logical isolation between diverging paths (e.g., separate travel plans). This design mirrors human cognitive separation, ensuring the model maintains a clear, distraction-free train of thought.

At each turn $t+1$, given:
\vspace{-8pt}
\begin{itemize}
    \item A structured dialogue history represented as a forest, $H_t = F_t$.
    \vspace{-10pt}
    \item The current state $S_t = \left( H_t, T_{\text{act}}, B_{\text{act}}, n_{\text{cur}} \right)$, which includes the history, the active topic tree, the active branch, and the current node.
    \vspace{-10pt}
    \item The new user query $q_{t+1}$.
\end{itemize}
The objective is to learn a policy $\pi$ that comprises two key functions: a context selection function, $f_{\text{select}}$, and a response generation function, $f_{\text{gen}}$:
\vspace{-5pt}
\[
C_{t+1} = f_{select}(q_{t+1}, S_t)
\]
\[
r_{t+1} = f_{gen}(q_{t+1}, C_{t+1})
\]
Here, $C_{t+1}$ represents a highly relevant context subset, which is dynamically selected and constructed from the structured history $H_t$. The ultimate goal is to maximize the task completion rate while minimizing the token footprint of the selected context $C_{t+1}$, thereby achieving efficient context utilization without compromising conversational coherence or task-oriented performance.

\begin{figure*}[t]
    \centering
    \includegraphics[width=0.95\textwidth]{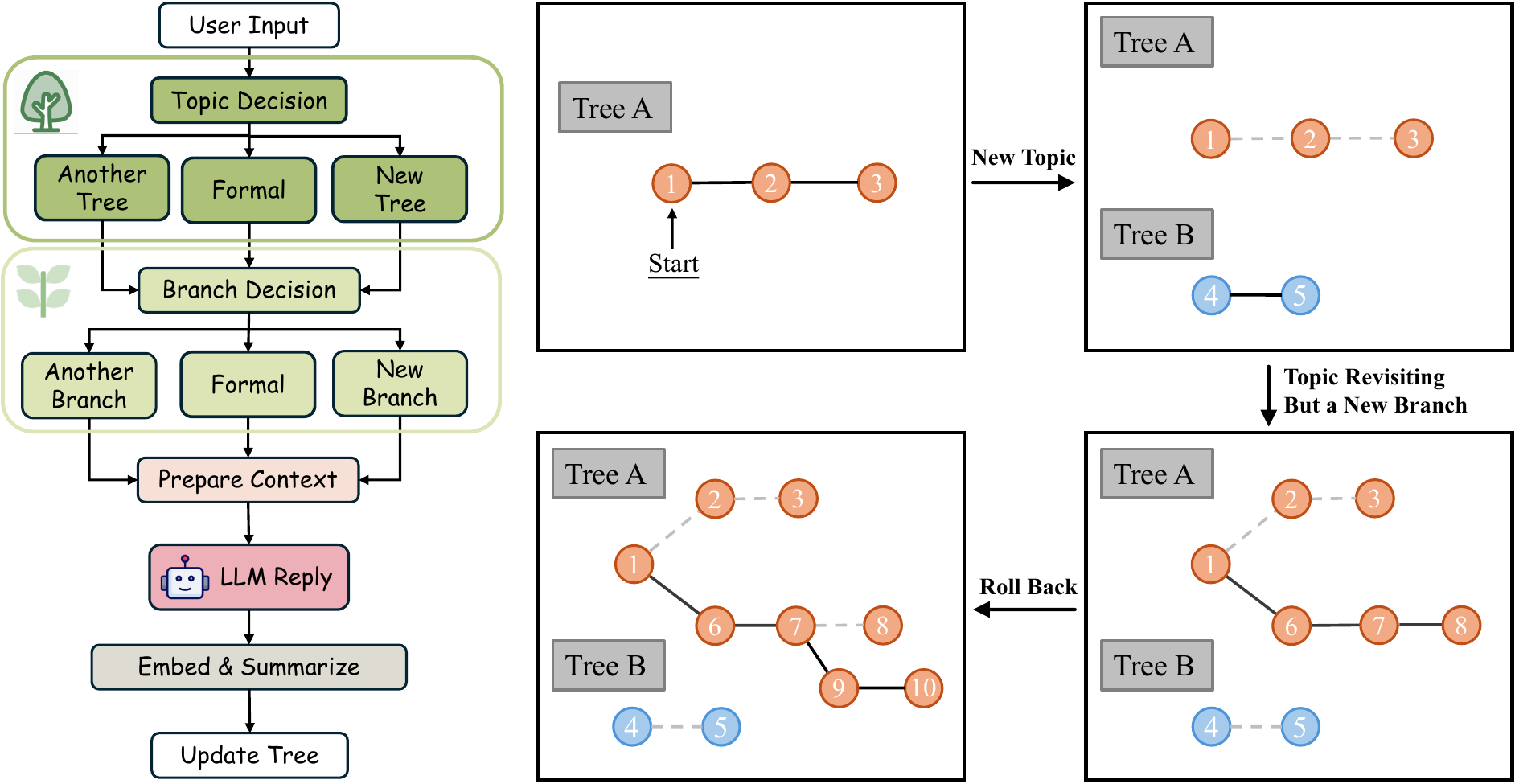}
    \caption{An overview of the Context-Agent framework. It illustrates the dynamic evolution of a multi-turn dialogue represented as a forest of topic trees, with branches indicating sub-dialogue paths. The number in each node represents the turn number in the conversation. Solid edges represent the active path, while dashed edges indicate inactive paths.}
    \label{fig:main-pic}
\end{figure*}

\subsection{Core Components}
\paragraph{\textbf{Node}} The smallest unit of a conversation is a node $n$, which represents the content of a round of dialogue between the user and the model. Each node is defined as a tuple:
$$n = (c, v, p, \beta, s_i)$$
where $c$ is the content of the current conversation round, $v \in \mathbb{R}^d$ is its $d$-dimensional text embedding, $p$ is the parent node's identifier (null for a root), $\beta$ is the branch identifier, and $s_i$ is a summary of the node's content. After each round, a summarization function $S_{node}$ converts the content $c_i$ into a summary $s_i = S_{node}(c_i)$, which is used for subsequent topic attribution and branch management.

\paragraph{Topic Tree} An independent topic is represented by a topic tree $T$. It is a directed acyclic graph, $T = (N, E)$. Here, $N=\{n_1, n_2, \ldots, n_k\}$ is the set of all nodes under this topic, and $E = \{(n_i, n_j) \mid p(n_j) = n_i\}$ is the set of directed edges between nodes, representing the inheritance relationship of the conversation. The first dialogue round of a new topic is set as the root node, whose parent node is null, of the topic tree.
\paragraph{Branch} Within the same topic tree $T$, a branch $B$ is a relatively independent dialogue path that starts from a branching point but still remains under the same topic. It is defined as an ordered sequence of nodes $B = \langle n_1, n_2, \ldots, n_k \rangle$, where any two adjacent nodes $(n_i, n_{i+1})$ in the sequence satisfy $p(n_{i+1}) = n_i$. All nodes within the same branch share the same branch identifier $\beta$.
\paragraph{Conversation History} The complete history $H$ of a multi-turn conversation is represented as a forest $F$ consisting of multiple topic trees, i.e., $H = F = \{T_1, T_2, \ldots, T_m\}$.

\subsection{State Transition}
The conversational state at turn $t$ is defined as $S_t = \left( H_t, T_{act}, B_{act}, n_{cur} \right)$, which includes the history, the active topic tree, the active branch, and the current node. The conversation evolves through state transitions driven by new user queries. Upon receiving a new query, the system analyzes it to determine the topic and manage branches, updating the state accordingly. This process involves the following steps:

\vspace{-7pt}
\begin{itemize}    
\item \textbf{Step0: Initialization} \quad Initialize the first topic tree $T_1$ as the active tree $T_{act}$. Define an aggregation function $S$ to summarize branches or trees by concatenating their constituent node summaries (e.g., $S(B) = \text{Concat}(s_1, \ldots, s_k)$).
    
\item \textbf{Step1: Topic Decision} \quad Given query $q_{t+1}$, a lightweight model $\Psi$ determines the action $a_{\text{topic}}$ and target tree $T_{\text{target}}$ using existing tree summaries:
    \[
        (a_{\text{topic}}, T_{\text{target}}) = \Psi(q_{t+1}, \{S(T_i)\})
    \]
    $T_{\text{act}}$ is updated to $T_{\text{target}}$. Actions include:
    \begin{itemize}[noitemsep]
        \item \textbf{CREATE\_TOPIC}: Start a new topic tree.
        \item \textbf{SWITCH\_TOPIC}: Switch to an existing tree.
        \item \textbf{CONTINUE}: Stay in the current tree.
    \end{itemize}

\item \textbf{Step2: Fork Point Identification} For a new query $q_{t+1}$, the system first computes its embedding vector $v_{q,t+1} = \epsilon(q_{t+1})$ using the embedding function $\epsilon : C \rightarrow \mathbb{R}^d$. Then, among all nodes in the active topic tree $T_{act}$, it identifies the node most semantically relevant to $q_{t+1}$ as the potential fork point. This is achieved by maximizing the similarity function $Sim(v_{q,t}, v_i)$:
    \[
    n_{\textit{fork}}^* = \arg \max_{n_i \in N_{\textit{act}}} \text{Sim}(v_{q,t+1}, v_i)
    \]
    
\item \textbf{Step3: Branch Decision} \quad Branch decision employs a two-stage ``heuristic filtering + model decision'' approach. First, a heuristic function $H_{\text{filter}}$ quickly determines if a complex decision is needed. Specifically, $H_{\text{filter}}$ returns true if the most similar node $n_{\textit{fork}}^*$ found in Step 2 is sufficiently relevant and it either belongs to a different branch or is an ancestor of the current node.
    
    If $H_{\text{filter}}$ is true, a lightweight language model $\Phi$ determines the branch action $a_{\text{branch}}$ based on the query, current path, and retrieved nodes $R(q)$. Otherwise, the action defaults to \text{CONTINUE}.
    \vspace{-3pt}
    \[
    a_{\text{branch}} =
    \begin{cases} 
    \Phi(q_{t+1}, \text{Path}(n_{\text{cur}}), R(q_{t+1})) \hspace{0.5em} H_{\text{filter}} \\
    \text{CONTINUE} \hspace{6.5em} \neg H_{\text{filter}}
    \end{cases}
    \]
    The possible actions are:
    \begin{itemize}[noitemsep]
        \item \textbf{CONTINUE}: Add a new node to the branch.
        \item \textbf{CREATE\_BRANCH}: Start a new branch from the fork point $n_{\textit{fork}}^*$.
        \item \textbf{SWITCH\_BRANCH}: Switch the active branch to the one containing $n_{\textit{fork}}^*$.
    \end{itemize}

\item \textbf{Step4: Context Construction} \quad The final context $C_{t+1}$ is constructed by combining the full dialogue of the current active path with summaries of inactive branches and topics. This provides focused, relevant information while maintaining a broad overview of the entire conversation. The context is formed as:
    \begin{equation*}
    \begin{gathered}
    C_{t+1} = \text{Concat}\bigl(\{ c_i \mid n_i \in \text{Path}(n_{\text{cur}}, T_{\text{act}}) \}\bigr) \\
    \bigoplus_{\substack{B_j \in T_{\text{act}}, \\ B_j \neq B_{\text{act}}}} S(B_j) \bigoplus_{\substack{T_k \in H_t, \\ T_k \neq T_{\text{act}}}} S(T_k)
    \end{gathered}
    \end{equation*}
    This structured context includes: (1) The complete dialogue history of the current active path. (2) Summaries of all other branches within the active topic tree. (3) Summaries of all other topic trees in the conversation history.
\end{itemize}

\section{Non-linear Task Multiturn Dialogue (NTM) Benchmark}

\label{benchmark}
Existing multi-turn datasets typically feature short (<10 turns), linear contexts \citep{DBLP:conf/acl/DeshpandeSMJHLK25, DBLP:conf/emnlp/KwanZJWLS00W24, DBLP:conf/acl/BaiLBHLZLSG0O24}, failing to capture the complexity of dynamic topic shifts essential for evaluating long-range reasoning. To bridge this gap, we introduce the Non-linear Task Multiturn Dialogue (NTM) benchmark.

\subsection{Data Creation}
NTM comprises a collection of dialogues focused on two domains: daily life planning and coding support. The dataset was constructed using state-of-the-art LLMs leveraging few-shot prompting to generate initial dialogues. Subsequently, each dialogue underwent a rigorous process of manual review, polishing, and filtering by human annotators to ensure high quality and task complexity.

\begin{figure*}[t]
    \centering
    \includegraphics[width=0.98\textwidth]{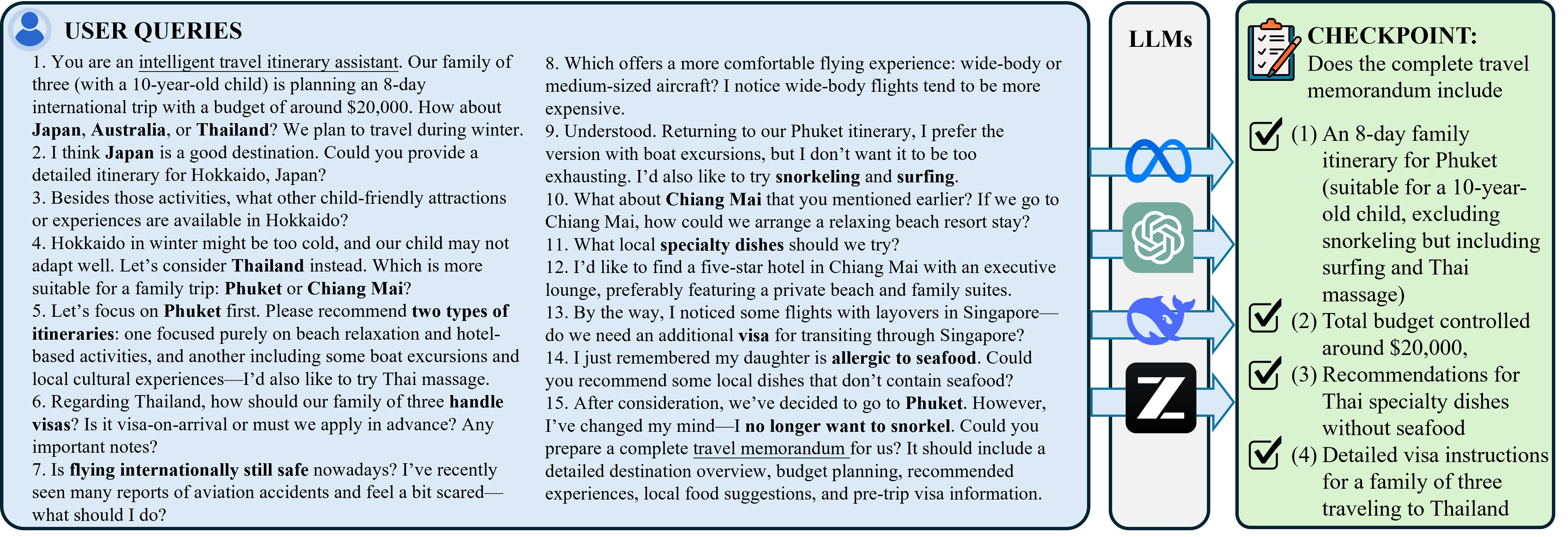}
    \caption{A 15-turn NTM dialogue example on trip planning, featuring topic shifts and instruction refinements. The right panel lists checkpoint questions for objective task completion evaluation. See Appendix \ref{appendix:benchmark} for details.}
    \label{fig:benchmark}
\end{figure*}

Crucially, NTM dialogues focus on two significant aspects: Topic shifts and Instruction Refinement, which are common in real-world conversations but often overlooked in existing datasets. 
\begin{itemize}
    \item \textbf{Topic Shifts}: Each dialogue is designed to include multiple topic shifts. These shifts are contextually relevant, reflecting how real conversations evolve. For example, a dialogue may start with planning a trip and then shift to discussing dietary preferences for the trip.
    
    \item \textbf{Instruction Refinement}: The dialogues also incorporate instances where users refine or change their instructions based on previous responses. This aspect tests the model's ability to adapt to evolving user needs and maintain coherence throughout the conversation.
\end{itemize}

This design ensures that NTM evaluates not just information recall, but a model's ability to maintain focus and adapt to a dynamically evolving conversational landscape.

\subsection{Key Characteristics}
NTM is distinguished by the following features:
\vspace{-6pt}
\begin{itemize}
    \item \textbf{Extended Dialogue Length}: The dataset includes a total of 405 dialogues with about 6900 turns, covering 10, 15, 20, and 25 rounds of conversations, which provide a clear measure of model scalability as context grows.
    \vspace{-5pt}
    \item \textbf{Topic Dynamics}: Each dialogue contains multiple topic shifts and instruction refinements, challenging models to maintain coherence and relevance in a non-linear conversational flow.
    \vspace{-5pt}
    \item \textbf{Task-Oriented Focus}: Every dialogue culminates in a clear task that requires accurate information synthesis from the preceding conversation, enabling objective evaluation through task completion metrics.
\end{itemize}

\subsection{Evaluation Metrics}
We evaluate the performance from 2 perspectives: task completion accuracy and token efficiency.
\begin{itemize}
    \item \textbf{Task Completion Rate (TCR)}: Our primary metric for task success. Each task in the NTM benchmark is decomposed into at least three verifiable checkpoints(a yes/no decision). TCR is the average completion rate across these checkpoints, providing a robust measure of task fulfillment. This annotated metric provides a more robust and interpretable measure of a model's true task-fulfillment capabilities compared to relying solely on scores from a judge LLM.

    \item \textbf{Average Context Tokens (ACT)}: Measures the average number of context tokens used per turn. It quantifies context efficiency, with lower values indicating better performance, which is crucial for managing long dialogues under token and cost constraints.
\end{itemize}

\subsection{Comparison with Existing Datasets}
Table~\ref{tab:dataset_comparison} compares NTM with existing datasets. NTM is distinguished by significantly longer turn counts and unique non-linear evolution, offering a more rigorous benchmark for complex dialogue evaluation.
\begin{table}[h]
    \centering
    \small
    \renewcommand{\arraystretch}{1.25}
    \begin{tabular}{lcccc}
    \hline
    \textbf{Dataset} &
    \begin{tabular}[c]{@{}c@{}}\textbf{Avg.}\\\textbf{Turns}\end{tabular} &
    \begin{tabular}[c]{@{}c@{}}\textbf{Max}\\\textbf{Turns}\end{tabular} &
    \begin{tabular}[c]{@{}c@{}}\textbf{Total}\\\textbf{turns}\end{tabular} &
    \begin{tabular}[c]{@{}c@{}}\textbf{Non-linear}\\\textbf{Evolution}\end{tabular} \\
    \hline
    Multichallenge & 5 & 10 & 1365 & No \\
    MT-Eval & 7 & 14 & 1170 & No \\
    MT-Bench-101 & 3 & 7 & 4208 & No \\
    \textbf{NTM (Ours)} & \textbf{17} & \textbf{27} & \textbf{6931} & \textbf{Yes} \\
    \hline
    \end{tabular}
    \caption{Comparison of NTM with existing multi-turn dialogue datasets.}
    \label{tab:dataset_comparison}
\end{table}

\section{Experimental Setup}
\label{experiments}
We conduct a comprehensive evaluation to assess Context-Agent's efficacy in managing long-form, non-linear dialogues, specifically examining its performance against baselines on complex tasks, its improvement in token efficiency relative to task success, and the distinct contributions of the tree-structured representation and retrieval mechanism.

\subsection{Evaluation Benchmarks}
A significant challenge in evaluating long-turn conversational models is the lack of suitable benchmarks. Existing datasets typically feature short, linear dialogues that do not adequately test a model's ability to handle complex, evolving conversations. And the most important reason is that their context offered to the model is usually a fixed-length linear sequence, which cannot reflect the advantages of our Context-Agent in managing non-linear dialogue history. Therefore, all models are evaluated on our newly proposed Non-linear Task Multi-turn Eval (NTM) benchmark.

To evaluate the generalizability of our method on public datasets, we selected TopiOCQA \citep{adlakha2022topiocqa} due to its rich topic shifts, which align well with our focus on non-linear dialogue management. We made appropriate adjustments to the dataset to facilitate testing within our framework, reporting Exact Match (EM) and F1 scores on the validation set.

\subsection{Baseline Methods}
We benchmark our Context-Agent framework against mainstream context management methods, which can be categorized into three groups:
\begin{itemize}
    \item \textbf{Full History Concatenation (Full-History)}: This method involves concatenating the entire dialogue history as input to the model. While it provides complete context, it is computationally expensive and often impractical for long conversations due to token limits.
    \vspace{-2pt}
    \item \textbf{Truncation (Truncation)}: This approach retains only the most recent $k$ turns of the conversation, discarding earlier context. It is efficient but risks losing important information from previous dialogue turns. In our experiments, we set $k=4$.
\end{itemize}

\begin{table}[h]
    \centering
    \small
    \begin{tabular}{lcc}
    \hline
    \textbf{Model} & \textbf{Open Source} & \textbf{Context Window} \\
    \hline
    GPT-4.1 & $\times$ & 1000k \\
    DeepSeek-V3 & $\checkmark$ & 64k \\
    GLM-4-Plus & $\times$ & 128k \\
    Llama 3.1-70B & $\checkmark$ & 128k \\
    \hline
    \end{tabular}
    \caption{Details of the LLMs used}
    \label{tab:model_details}
\end{table}

To ensure a comprehensive evaluation of our Context-Agent across different models, we conducted experiments on four recent and diverse LLMs: \texttt{GPT-4.1} \citep{OpenAI2025}, \texttt{DeepSeek-V3} \citep{liu2024deepseek}, \texttt{GLM-4-Plus} \citep{glm2024chatglm}, and \texttt{Llama 3.1-70B} \citep{grattafiori2024llama}. This selection includes both open- and closed-source models with varying context window sizes. For fairness and efficiency, all evaluations were performed with reasoning-disabled settings.

\subsection{Implementation Details}

To balance processing efficiency and accuracy, we employ \texttt{gemma3-12B} \citep{team2025gemma} for decision-making and \texttt{gemma3-4B} for summary generation. For dialogue context encoding, we use \texttt{Qwen3-Embedding-0.6B} \citep{yang2025qwen3}. All experiments were conducted with an NVIDIA A100 40GB GPU. For evaluation, we adopt a triangulated protocol combining human annotators and Judge LLMs (\texttt{GPT-5} and \texttt{Gemini-2.5-Pro}). For more details, please refer to Appendix \ref{appendix:implementation}. 

\section{Results and Analysis}
\label{results}

\subsection{Main Results}
The main results of our experiments are summarized in Table \ref{tab:main_results}. Across all four LLMs, our Context-Agent consistently outperforms the Truncation method by a significant margin in terms of Task Completion Rate (TCR). Notably, our method not only recovers the performance loss caused by truncation but also surpasses the Full-History method across the board. Specifically, it achieves relative TCR improvements of 3.4\%, 7.8\%, 8.1\%, and 9.7\% on \texttt{GPT-4.1}, \texttt{DeepSeek-V3}, \texttt{GLM-4-Plus}, and \texttt{Llama 3.1-70B}, respectively. Even for \texttt{GPT-4.1}, which possesses a massive context window, Context-Agent achieves a score of 88.9\%, outperforming the Full-History score of 86.0\%. This suggests that structured context management effectively filters noise that can distract even the most capable models. Furthermore, Context-Agent demonstrates superior efficiency, reducing the Average Context Tokens (ACT) by approximately 45\% to 52\% compared to the Full-History approach. This dual advantage of higher accuracy and lower token consumption underscores the efficacy of the Context-Agent.

Table \ref{tab:topiocqa} demonstrates Context-Agent's robust generalization on TopiOCQA. It outperforms Full-History in accuracy (EM/F1) while using only $\sim$57\% of the context tokens. This efficiency stems from the tree-structured memory, which isolates the active topic to minimize noise without losing necessary context.

\begin{table*}[t]
    \centering
    \small
    \setlength{\tabcolsep}{3pt}
    \renewcommand{\arraystretch}{1.1}
    \begin{tabular*}{\textwidth}{@{\extracolsep{\fill}} l l c c cccc c}
    \toprule
    \multirow{2}{*}{\textbf{Model}} & \multirow{2}{*}{\textbf{Method}} & \multirow{2}{*}{\textbf{TCR (\%) $\Uparrow$}} & \multirow{2}{*}{\makecell{\textbf{TCR} \\ \textbf{Gain(\%)}}} & \multicolumn{4}{c}{\textbf{ACT $\Downarrow$}} & \multirow{2}{*}{\makecell{\textbf{ACT} \\ \textbf{Drop (\%)}}} \\
    \cmidrule{5-8}
    & & & & 10-turn & 15-turn & 20-turn & 25-turn & \\
    \midrule
    GPT-4.1
    & Full-History    & 86.0 & -- & 4070 & 6382 & 9535 & 12803 & -- \\
    & Truncation      & 55.2 & -35.8 & 1839 & 2378 & 2981 & 3142 & -- \\
    & \textbf{Context-Agent} & \textbf{88.9} & \textbf{+3.4} & 2108 & 2894 & 4137 & 6227 & \textbf{-52.3} \\
    \midrule
    DeepSeek-V3
    & Full-History    & 64.3 & -- & 3540 & 5428 & 7805 & 10693 & -- \\
    & Truncation      & 42.8 & -33.4 & 1732 & 2088 & 2535 & 2883 & -- \\
    & \textbf{Context-Agent} & \textbf{69.3} & \textbf{+7.8} & 1914 & 2873 & 4110 & 6014 & \textbf{-46.0} \\
    \midrule
    GLM-4-Plus
    & Full-History    & 71.5 & -- & 4130 & 6996 & 9403 & 11782 & -- \\
    & Truncation      & 45.1 & -36.9 & 2890 & 3479 & 3783 & 4674 & -- \\
    & \textbf{Context-Agent} & \textbf{77.3} & \textbf{+8.1} & 1954 & 3027 & 4695 & 7032 & \textbf{-49.9} \\
    \midrule
    Llama 3.1-70B
    & Full-History    & 65.1 & -- & 3540 & 5183 & 7189 & 8994 & -- \\
    & Truncation      & 44.0 & -32.4 & 1689 & 1898 & 2435 & 2860 & -- \\
    & \textbf{Context-Agent} & \textbf{71.4} & \textbf{+9.7} & 2075 & 2738 & 3843 & 4780 & \textbf{-45.5} \\
    \bottomrule
    \end{tabular*}
    \caption{\textbf{Main Results on Context Management Efficiency and Effectiveness.} 
    Performance on our proposed \textbf{NTM Benchmark} (Task-Oriented) across varying dialogue lengths.
    \textbf{TCR}: Task Completion Rate; \textbf{ACT}: Average Context Tokens. Context-Agent consistently outperforms baselines.}
    \label{tab:main_results}
\end{table*}

\begin{table}[t]
    \centering
    \small
    \setlength{\tabcolsep}{5pt}
    \renewcommand{\arraystretch}{1.1}
    \begin{tabular}{lccc}
    \toprule
    \textbf{Method} & \textbf{EM(Exact Match)} & \textbf{F1 Score} & \textbf{ACT} \\
    \midrule
    Full-History & 13.3 & 25.2 & 4261 \\
    Truncation & 7.1 & 12.8 & 1703 \\
    \textbf{Context-Agent} & \textbf{16.2} & \textbf{28.9} & \textbf{2435} \\
    \bottomrule
    \end{tabular}
    \caption{Result of \texttt{Llama 3.1-70B} on TopiOCQA.}
    \label{tab:topiocqa}
    \vspace{0.1cm}
    \begin{minipage}{\linewidth}
    \scriptsize
    \end{minipage}
\end{table}

\begin{figure}[h]
    \centering
    \makebox[\linewidth][c]{
    \subfigure[]{
        \includegraphics[width=0.63\linewidth]{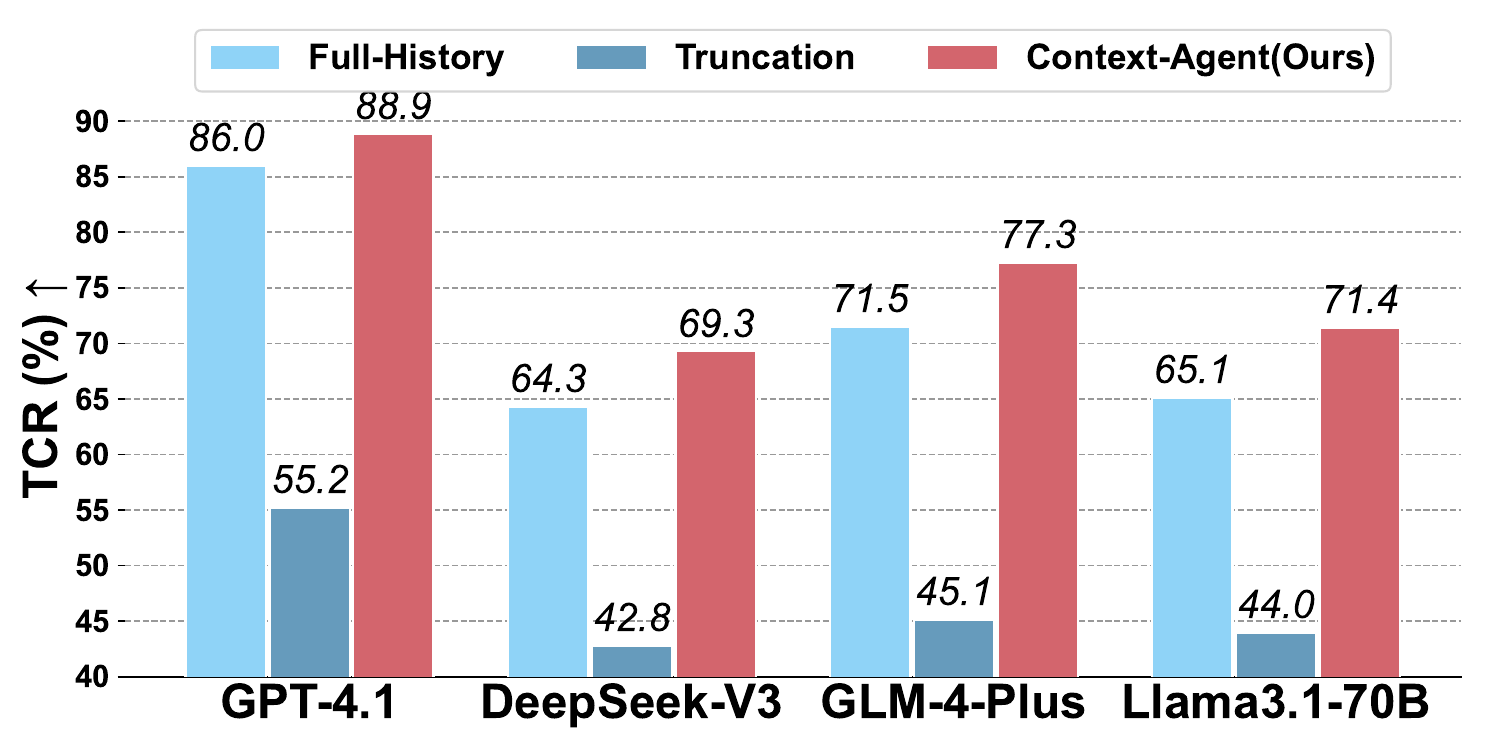}}
    \hspace{0.02\linewidth}
    \subfigure[]{
        \includegraphics[width=0.33\linewidth]{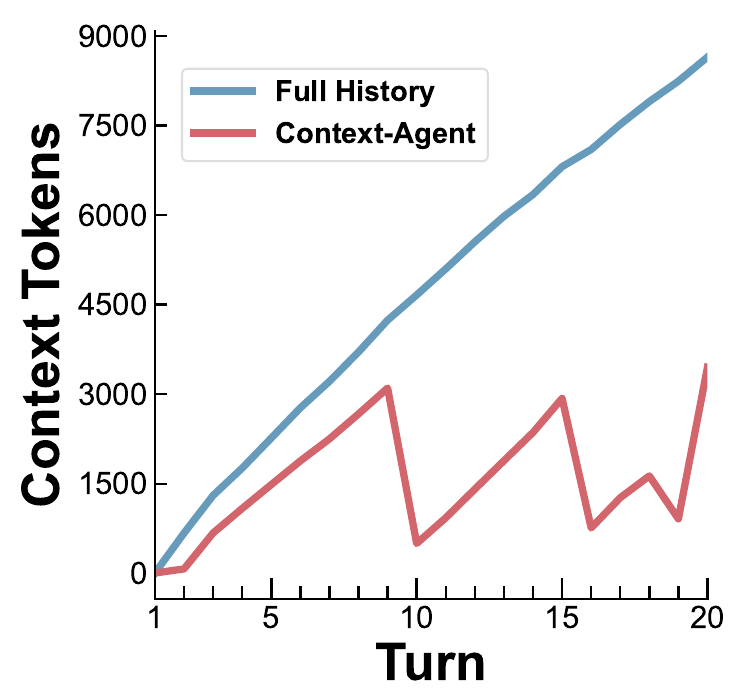}}
    }
    \caption{(a) TCR comparison across different methods and models. (b) A typical example of context tokens change trend in a 20-turn dialogue.}
    \label{fig:main-result-ab}
\end{figure}

\begin{figure}[h]
    \centering
    \includegraphics[width=0.95\linewidth]{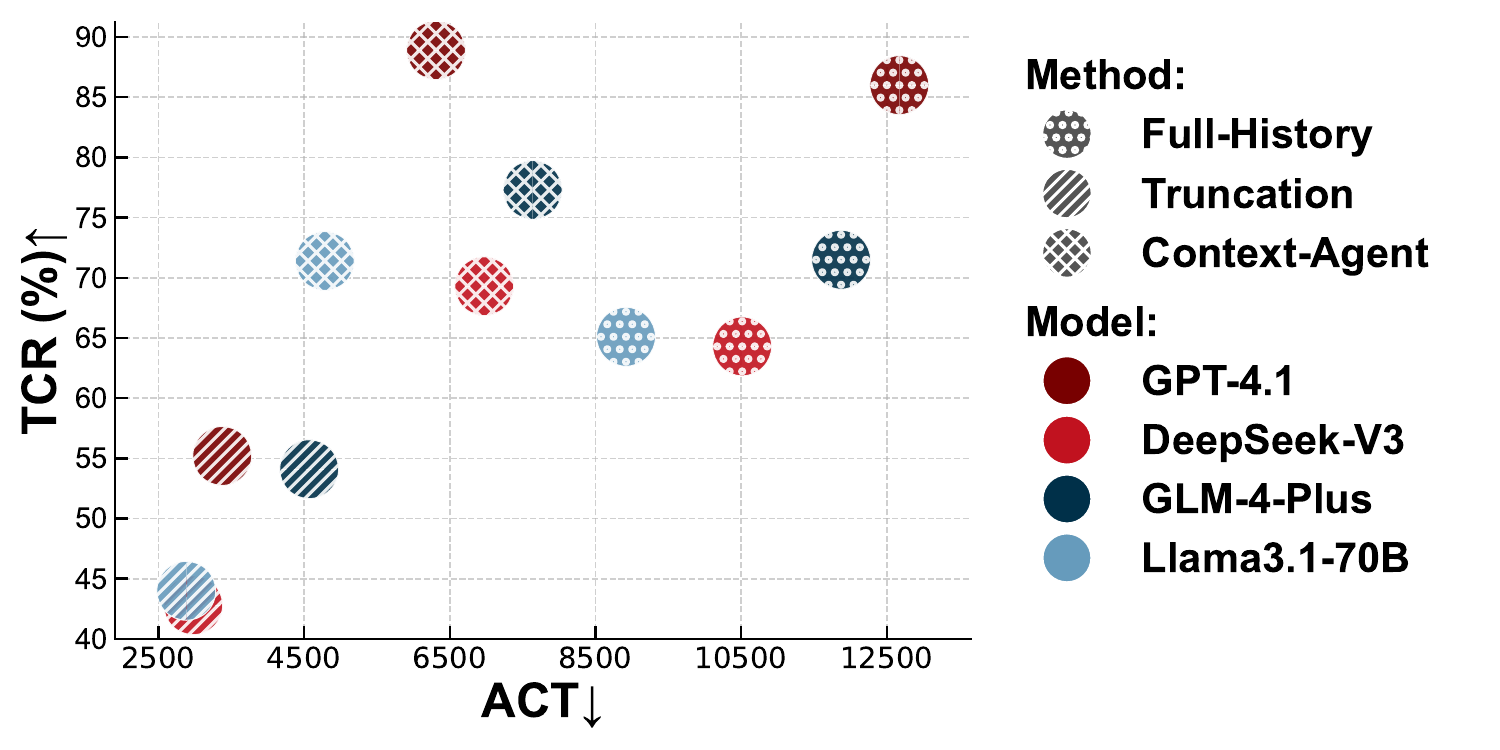}
    \caption{Trade-off between TCR and ACT, where the ideal point is the top-left corner (high TCR, low ACT).}
    \label{fig:main-result-c}
\end{figure}

Another notable observation is that though another 3 open-source models (\texttt{DeepSeek-V3}, \texttt{GLM-4-Plus}, and \texttt{Llama 3.1-70B}) still have considerable context windows (64k or 128k tokens), and the total context length of our NTM benchmark is lower than these limits, their TCR scores with Full-History are still significantly lower than that of \texttt{GPT-4.1}. This indicates that merely having a large context window does not guarantee effective utilization of context, especially in complex, non-linear dialogues. Our Context-Agent has demonstrated its ability to effectively manage and utilize context, leading to substantial performance gains.

From these results, we have several key insights:

\begin{itemize}
    \item \textbf{Effectiveness of Context-Agent}: The consistent TCR improvements across different models and dialogue lengths demonstrate that Context-Agent effectively manages context in complex, long-horizon dialogues. It not only recovers the performance lost due to truncation but also surpasses the full-history approach in most cases.
    
    \item \textbf{Token Efficiency}: The significant reductions in ACT indicate that Context-Agent is highly efficient in utilizing context. By intelligently selecting relevant information through its tree structure and RAG mechanism, it minimizes unnecessary token usage while still providing sufficient context for accurate responses.
    
    \item \textbf{Robustness Across Models}: The performance gains observed across a diverse set of LLMs, including both open-source and closed-source models with varying context window sizes, highlight the robustness and generalizability of the Context-Agent framework.
\end{itemize}

\subsection{Ablation Studies}
To isolate component contributions, we conducted an ablation study (Table \ref{tab:ablation_study}). We evaluated two variants: (1) \textbf{w/o Tree}, which applies RAG to a flattened linear history (retrieving $k \in \{3, 5\}$ turns), and (2) \textbf{w/o RAG}, which relies solely on heuristics for branch decisions without semantic retrieval.

\begin{table}[h]
    \centering
    \small
    \setlength{\tabcolsep}{6pt}
    \renewcommand{\arraystretch}{1.25}
    \begin{tabular}{l c c}
    \hline
    \textbf{Method} & \textbf{TCR (\%)} & \textbf{TCR Drop (\%)} \\
    \hline
    Full-History & 64.3 & - \\
    w/o Tree & 41.5 & -35.5\\
    w/o RAG & 45.3 & -29.5 \\
    Context-Agent & 69.3 & +7.8 \\
    \hline
    \end{tabular}
    \caption{Ablation study results on DeepSeek-V3.}
    \label{tab:ablation_study}
\end{table}

Results indicate that both components are essential. Removing the tree structure (\textbf{w/o Tree}) leads to a 35.5\% TCR drop, confirming that linear retrieval captures semantic similarity but fails to maintain the logical flow necessary for effective context selection. Similarly, removing the retriever (\textbf{w/o RAG}) results in a 29.5\% drop, showing that heuristics alone are insufficient for accurate fork point identification.

\section{Conclusion}
\label{conclusion}
In this paper, we addressed the critical limitation of conventional linear context management in handling the non-linear flow of multi-turn dialogues. We introduced Context-Agent, a novel framework that represents dialogue history as a dynamic tree structure, augmented by a retrieval mechanism. This approach successfully models the hierarchical and branching nature of human conversations, enabling effective navigation of complex interactions involving topic shifts and refinements. Our extensive experiments on the newly proposed NTM benchmark demonstrate that Context-Agent consistently outperforms traditional context management methods across various LLMs, achieving significant improvements in task completion rates while drastically reducing token usage. Ablation studies confirm the critical contributions of both the tree structure and RAG components to the overall performance. Our work underscores the potential of structured context management and offers a promising direction for developing more robust and efficient dialogue systems capable of handling long-horizon, dynamic conversations.

\section*{Limitations}
Current implementation relies on lightweight models for topic and branch decisions, whose performance may vary with model choice and prompting strategies. While our experiments show consistent gains across multiple backbones, further optimizing or learning these decision modules end-to-end could potentially yield additional improvements.

\bibliography{main}

\clearpage

\appendix

\section{Appendix}

\subsection{Reproductivity Statement}
To facilitate future research, we will fully \textbf{open-source} the Context-Agent, the NTM benchmark dataset, and all relevant experimental scripts upon the acceptance of this paper. Relevant code and data are currently attached for review.

\subsection{Implementation Details}
\label{appendix:implementation}

\textbf{Prompt Format:} All models receive the same system prompt instructing them. No chain-of-thought or explicit instruction tuning is applied to ensure fair comparison. More details are in Appendix \ref{appendix:prompts}.

\textbf{Local Models:} To balance processing efficiency and accuracy, the Context-Agent's internal modules utilize lightweight local models. Specifically, we employ \texttt{gemma3-12B} \citep{team2025gemma} for decision-making and \texttt{gemma3-4B} for summary generation. For dialogue context encoding, we use \texttt{Qwen3-Embedding-0.6B} \citep{yang2025qwen3}, a lightweight, high-performance embedding model. Based on empirical tuning with these models, the similarity threshold $\theta_{\text{sim}}$ was set to 0.6. All experiments were conducted with an NVIDIA A100 40GB GPU.

\textbf{Evaluation Protocol:} To ensure both scalability and human-aligned judgment, we adopt a triangulated evaluation protocol combining human annotators and two state-of-the-art Judge LLMs: \texttt{GPT-5} \citep{gpt52025} and \texttt{Gemini-2.5-Pro} \citep{comanici2025gemini}. We compute Cohen’s $\kappa$ \citep{cohen1960coefficient} between Judge LLM and human labels. The result shows that the Cohen's $\kappa$ is as high as 0.96, indicating strong agreement and validating the reliability of our evaluation approach.

\subsection{Context-Agent Latency and Trade-off Analysis}
Beyond token efficiency, we analyzed the end-to-end response latency to provide a complete picture of Context-Agent's practical performance. Our method's hybrid architecture involves several calls to local, lightweight language models for tasks such as branch decision-making and node summarization, which introduces time overhead compared to the baseline's single API call.

However, the latency of the full-context baseline is not constant; it degrades as the dialogue history grows and the token payload for the API call increases. This degradation partially offsets the inherent overhead of our method. To quantify this trade-off, we measured the average response time on a single NVIDIA A100 40GB GPU for the 20-turn dialogue scenario. The following table summarizes the average response times:
\begin{table}[h]
    \centering
    \small
    \setlength{\tabcolsep}{6pt} 
    \renewcommand{\arraystretch}{1.25} 
    \begin{tabular}{l c c}
    \hline
    \textbf{Method} &
    \begin{tabular}[c]{@{}c@{}}\textbf{Average}\\\textbf{Response Time(s)}\\\end{tabular} & 
    \begin{tabular}[c]{@{}c@{}}\textbf{Relative}\\\textbf{Increase(\%)}\\\end{tabular} \\
    \hline
    Full-History & 12.5 & - \\
    Context-Agent & 13.5 & +8.0\% \\
    \hline
    \end{tabular}
    \caption{Average response time for different context management methods on a 20-turn dialogue.}
    \label{tab:latency_analysis}
\end{table}

Our experiments indicate that Context-Agent incurs a modest 8\% increase in average response time. We argue this represents a highly favorable trade-off, given the substantial improvements in token efficiency. It is important to note that these measurements were conducted on a single A100 40GB GPU. This latency overhead could likely be mitigated in a production environment through optimizations such as deploying on enterprise-grade hardware or utilizing lightweight models fine-tuned for the specific decision and summarization sub-tasks.

\subsection{the Detailed Algorithm of Context-Agent}
The complete algorithm of the Context-Agent framework is presented in Algorithm \ref{alg:context-agent}. It outlines the step-by-step process of managing dialogue context, including topic and branch management, node updates, and context construction.

\begin{algorithm}[h]
\small
\caption{Context-Agent Framework}
\label{alg:context-agent}
\begin{algorithmic}[1]
\Require Dialogue history $H_t$, User query $q_{t+1}$
\Ensure Constructed context $C_{t+1}$

\Statex \textbf{1. Topic and Branch Management}
\State $(a_{\text{topic}}, T_{\text{target}}) \gets \Psi(q_{t+1}, \{S(T_i)\}_{T_i \in H_t})$ \Comment{Topic decision}
\State Update $T_{\text{act}}, n_{\text{cur}}$ based on $a_{\text{topic}}$
\State $n_{\textit{fork}}^* \gets \arg \max_{n_i \in T_{\text{act}}} \text{Sim}(\epsilon(q_{t+1}), v_i)$ \Comment{Find fork point}
\If{$H_{\text{filter}}(n_{\textit{fork}}^*, n_{\text{cur}})$}
    \State $a_{\text{branch}} \gets \Phi(q_{t+1}, \text{Path}(n_{\text{cur}}), R(q_{t+1}))$ \Comment{Branch decision}
\Else
    \State $a_{\text{branch}} \gets \text{CONTINUE}$
\EndIf
\State Update $B_{\text{act}}, n_{\text{cur}}$ based on $a_{\text{branch}}$ and $n_{\textit{fork}}^*$

\Statex \textbf{2. Node Update}
\State Create new node $n_{\text{new}}$ as child of $n_{\text{cur}}$
\State $s_{\text{new}} \gets S_{\text{node}}(n_{\text{new}})$ \Comment{Summarize new node}
\State $n_{\text{cur}} \gets n_{\text{new}}$

\Statex \textbf{3. Context Construction}
\State $C_{\text{path}} \gets \{ c_i \mid n_i \in \text{Path}(n_{\text{cur}}) \}$ \Comment{Content of active path}
\State $C_{\text{inactive}} \gets \{ S(B_j) \mid B_j \neq B_{\text{act}} \} \cup \{ S(T_k) \mid T_k \neq T_{\text{act}} \}$ \Comment{Summaries of inactive parts}
\State $C_{t+1} \gets \text{Concat}(C_{\text{path}}, C_{\text{inactive}})$
\State \Return $C_{t+1}$
\end{algorithmic}
\end{algorithm}

\newpage
\subsection{Model Implementation Details}
\label{appendix:prompts}
This section provides the specific prompts used to guide the lightweight language models for decision-making and summarization within the Context-Agent framework.

\textbf{Prompt for Topic Decision} \quad
The following prompt is used to instruct the topic decision model $\Psi$ to analyze the user's query against the summaries of existing topic trees. The model must determine whether the query initiates a new topic, continues the current one, or switches to a previous one.
\begin{figure}[h]
    \centering
    \includegraphics[width=0.98\linewidth]{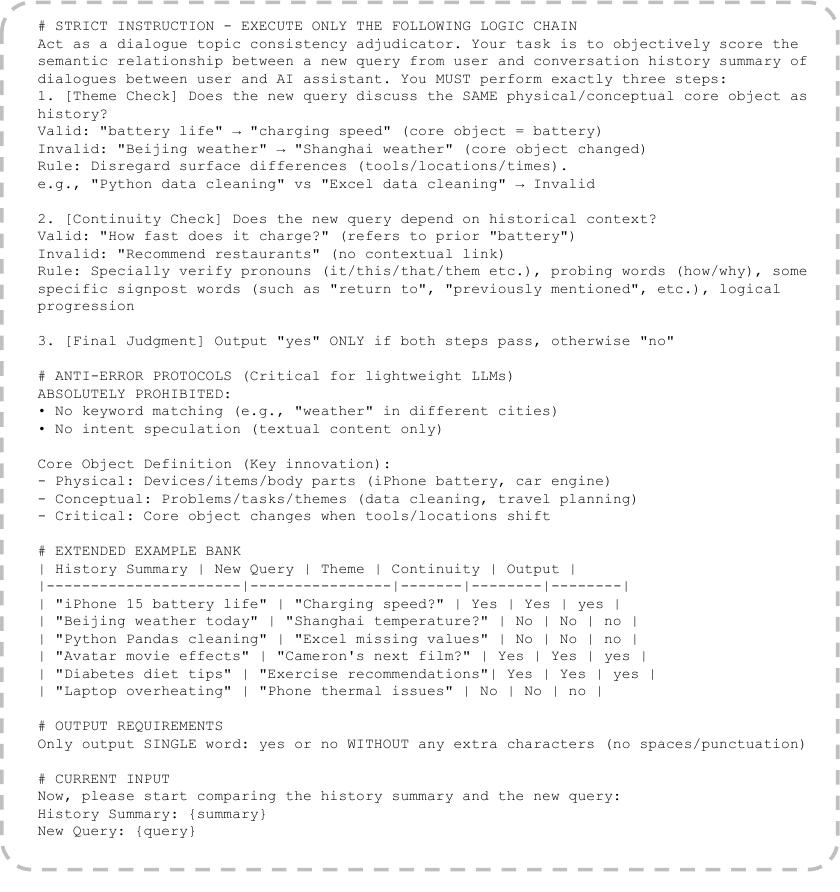}
    \vspace{-25pt}
    \caption*{}
    \label{fig:prompt-topic}
\end{figure}

\textbf{Prompt for Branch Decision} \quad
The branch decision model $\Phi$ is prompted to evaluate the user's query in the context of the current dialogue path and the most relevant historical nodes. The model must decide whether to continue the current branch, create a new branch, or switch to an existing one.
\begin{figure}[!h]
    \centering
    \includegraphics[width=0.98\linewidth]{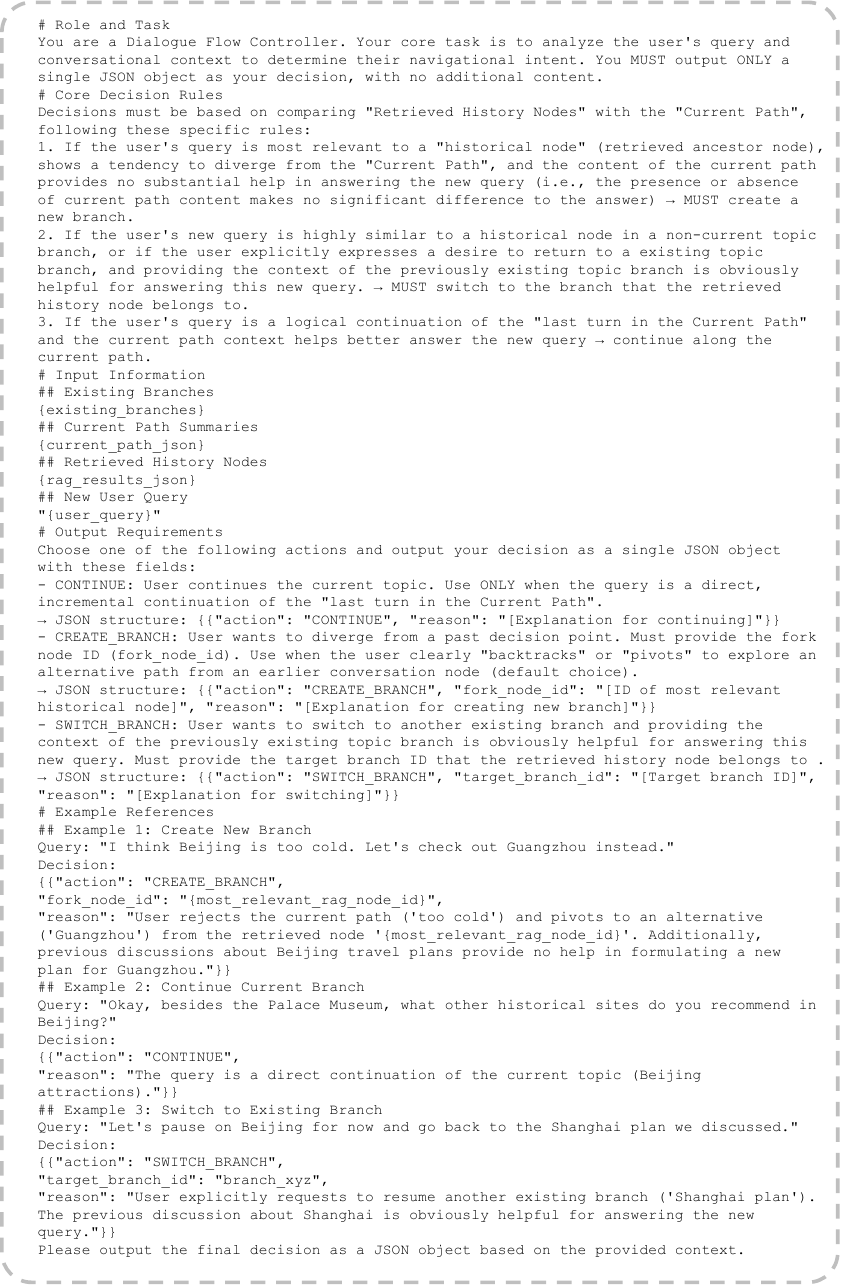}
    \vspace{-25pt}
    \caption*{}
    \label{fig:prompt-branch}
\end{figure}

\textbf{Prompt for Node Summarization} \quad
The node summarization model $S_{node}$ is prompted to generate concise summaries of dialogue nodes. The prompt emphasizes the need for brevity and relevance, ensuring that the summaries capture the essence of each node for effective context management.
\begin{figure}[!h]
    \centering
    \includegraphics[width=0.98\linewidth]{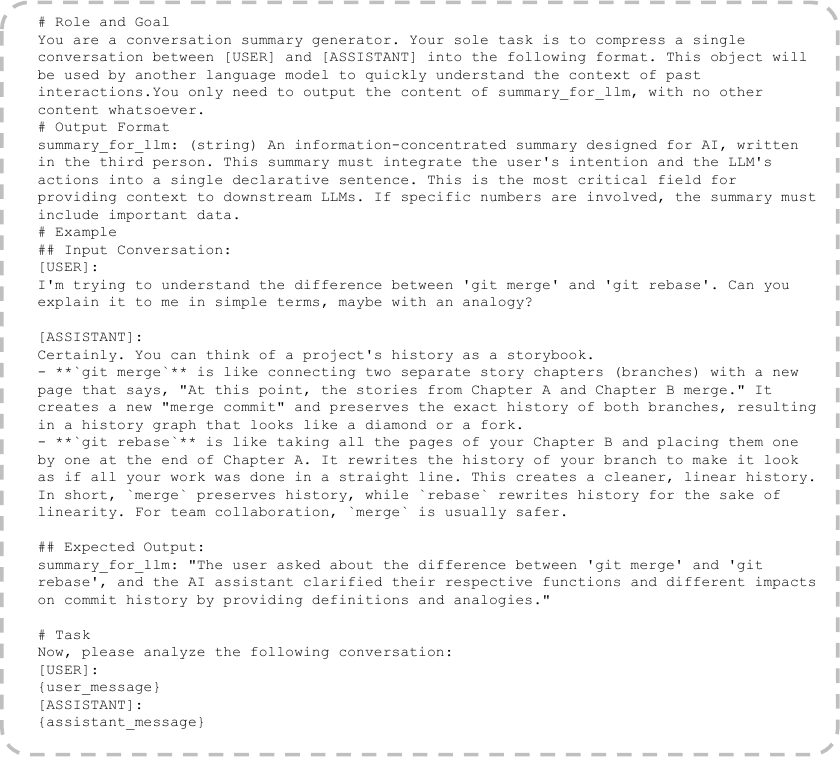}
    \vspace{-25pt}
    \caption*{}
    \label{fig:prompt-summarize}
\end{figure}

\subsection{NTM Benchmark Details}
\label{appendix:benchmark}
The Non-linear Task Multiturn Dialogue (NTM) benchmark is designed to evaluate the performance of dialogue systems in handling complex, multi-turn conversations with dynamic topic shifts and instruction refinements. Such dynamic context evaluation aligns with the growing need to assess agents in ever-changing environments, analogous to proactive experience-seeking in web agents \citep{zhang2026expseek}. Below are the details of the NTM benchmark.

\subsubsection{Human Annotation Guidelines}
To ensure the quality and consistency of the NTM benchmark, human annotators reviewed, polished, and filtered the generated dialogues based on the following primary criteria:
\vspace{-2pt}
\begin{itemize}
    \item \textbf{Coherence and Naturalness}: The dialogue must flow logically and feel natural, avoiding robotic or repetitive responses. Topic shifts, a key feature of the benchmark, must be contextually plausible and not feel abrupt or random. The overall conversation should mimic the ebb and flow of genuine human interaction, including clarifications, refinements, and relevant digressions.
    \item \textbf{Task Complexity}: Each dialogue must build towards a clear, non-trivial final task. Successfully completing this task should require the model to synthesize and integrate information scattered across multiple turns, including handling user refinements and instruction changes. Simple, single-turn information retrieval is insufficient; the task must test long-range reasoning and memory.
    \item \textbf{Clarity and Objectivity of Checkpoints}: To facilitate objective and reproducible evaluation, the final task must be decomposable into a set of clear, unambiguous, and verifiable checkpoints. Each checkpoint should correspond to a specific sub-goal of the user's final request and be answerable with a simple ``yes'' or ``no'', minimizing subjective judgment during evaluation.
\end{itemize}

\begin{figure*}[t!]
  \centering
  \includegraphics[width=0.9\textwidth]{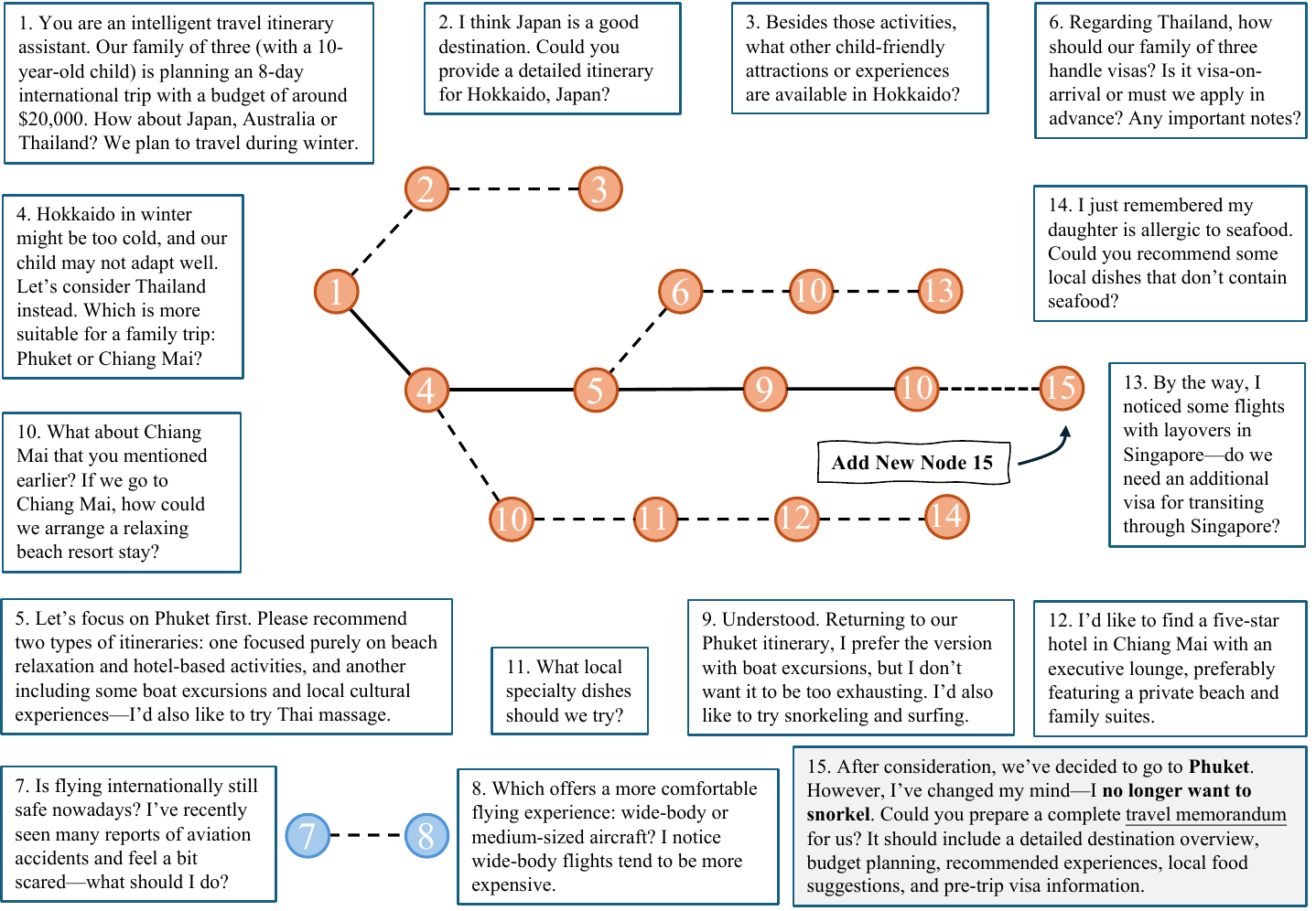}
  \caption{The topic tree structure corresponding to the dialogue example in Figure~\ref{fig:benchmark}. Each node represents a turn in the dialogue, with branches indicating topic shifts and refinements. The solid edges represent the active path, while the dashed edges represent inactive branches.}
  \label{fig:example_tree}
\end{figure*}

\subsubsection{The detailed topic trees} \quad
In the previous Figure ~\ref{fig:benchmark} in Section ~\ref{benchmark}, we provided a dialogue example. To more intuitively demonstrate the formation of the dialogue tree, we have visualized the dialogue example shown in Figure ~\ref{fig:benchmark} into a tree structure.

Showed in Figure ~\ref{fig:example_tree}, the dialogue starts with planning a family trip. In the first turn, the user introduces the plan and suggests several potential destinations, which sets a potential fork point for the future exploration of different destinations. Then the user and the assistant discuss the details of the Hokkaido itinerary, including child-friendly attractions. However, in turn 4, the user shifts the topic to Thailand due to concerns about the cold weather in Hokkaido. This shift is still within the topic of trip planning but introduces a new destination. And it is totally different from the previous discussing about Japan. The history of the first three turns is not so useful for the following discussion about Thailand. 

Therefore, the Context-Agent creates a new topic tree for Thailand, starting a new branch from turn 4. The user then explores two potential locations in Thailand: Phuket and Chiang Mai, requesting different types of itineraries and activities. This introduces another fork point at turn 5, where the user asks for two distinct itinerary options for Phuket.

In turn 7, the user raises a concern about the safety of international flights, which is totally different from the previous topic of trip planning. This prompts the Context-Agent to create another new topic tree for flight safety, starting a new tree from turn 7. The user and assistant discuss various aspects of flying, including aircraft types and comfort. 

Then in turn 9, the user returns to the Phuket itinerary, indicating a switch back to the previous topic tree about Thailand. The Context-Agent recognizes this and switches the active topic tree back to Thailand. The user continues to refine their preferences for the Phuket itinerary, expressing a desire for a more relaxing experience without snorkeling. Nevertheless, in turn 10, the user again shifts the focus to Chiang Mai, asking about arranging a beach resort stay there. This indicates another switch within the Thailand topic tree. And in turn 14, the user refines their food preferences due to a seafood allergy. Finally, in turn 15, the user makes a final decision to go to Phuket but changes their mind about snorkeling and requests a comprehensive travel memorandum that synthesizes all the discussed information, including destination overview, budget planning, recommended experiences, local food suggestions, and visa information.

\subsubsection{Example from ``Coding Support'' Domain}   \quad
This example illustrates a typical dialogue from the NTM benchmark's coding support domain, featuring topic shifts and instruction refinements.

\begin{figure*}[t]
    \centering
    \includegraphics[width=0.95\textwidth]{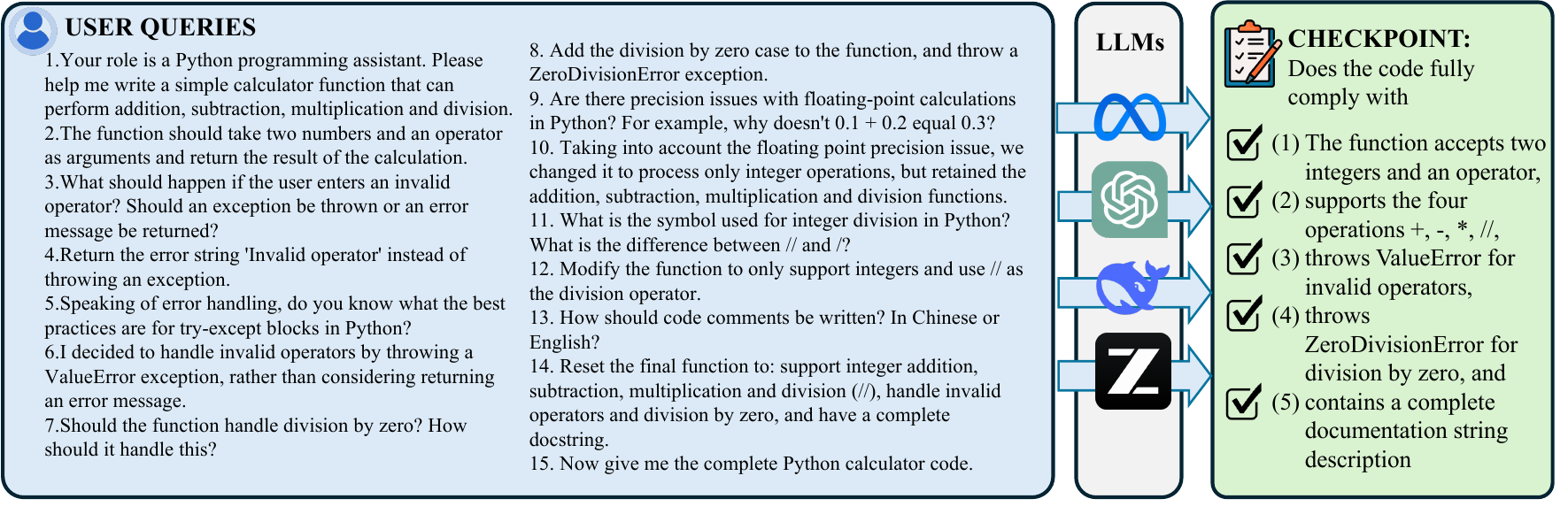}
    \caption{An example of a 15-turn dialogue from the NTM benchmark in the coding support domain. The dialogue features multiple topic shifts and instruction refinements, culminating in a clear task of generating a Python calculator function.}
    \label{fig:coding_example}
\end{figure*}

As shown in Figure~\ref{fig:coding_example}, the dialogue begins with a request for a basic calculator. The user iteratively refines the requirements—adding error handling and changing data types from floats to integers—while also digressing to discuss `try-except' best practices and commenting conventions. Finally, the user consolidates all refinements into a final request for the complete code. This example highlights the benchmark's focus on testing a model's ability to handle instruction changes, topic shifts, and integrate information from a non-linear dialogue history.

\end{document}